\pdfoutput=1

\documentclass[11pt]{article}

\usepackage{acl}

\usepackage{latexsym}
\usepackage{natbib}  
\usepackage{caption} 

\usepackage[utf8]{inputenc} 
\usepackage[T1]{fontenc}    
\usepackage{hyperref}       
\usepackage{url}            
\usepackage{booktabs}       
\usepackage{amsfonts}       
\usepackage{nicefrac}       
\usepackage{microtype}      
\usepackage{xcolor}         
\usepackage{verbatim}
\usepackage{bm}

\usepackage{times}
\usepackage{latexsym}
\usepackage[T1]{fontenc}
\usepackage[utf8]{inputenc}
\usepackage{microtype}
\usepackage{times}
\usepackage{natbib} 
\usepackage{caption} 
\usepackage{amsmath}
\usepackage{amsthm}

\usepackage{subfigure}
\usepackage{graphicx}
\usepackage{algorithm}
\usepackage{algorithmic}
\usepackage{parskip}
\usepackage{booktabs}
\usepackage{multirow}
\usepackage{amssymb}
\usepackage{hyperref}

\usepackage[normalem]{ulem}
\useunder{\uline}{\ul}{}

\title{Parameter-Efficient Tuning on Layer Normalization for Pre-trained Language Models}
 
\author{
Wang Qi$^1$,
Yu-Ping Ruan$^1$,
Yuan Zuo$^2$,
\and 
Taihao Li$^1$\thanks{$^*$The corresponding author.} \\
$^1$AI Research Institute, Zhejiang Lab, Hangzhou\\
$^2$Department of Information Systems, Beihang University \\
\texttt{\{qiwang, ypruan, lith\}@zhejianglab.com,
 zuoyuan@buaa.edu.cn}
}
 
\begin{document}

	\maketitle
	
	\begin{abstract}
 
 Given the magnitude of the current Pre-trained Language Models (PLMs), conventional fine-tuning becomes increasingly challenging, therefore parameter-efficient tuning is now the focus of cutting-edge research. For PLMs to accomplish transferability, prior techniques in this field added tunable adapters into Multi-Head Attention (MHA) or/and Feed-Forward Network (FFN) of Transformer blocks. However, the ability of Layer Normalization (LayerNorm) for parameter-efficient tuning is disregarded while being a crucial component of Transformer architecture. In this paper, we first propose LN-tuning, which is time-efficient and performs much better than baselines with less than 0.1\% tunable parameters by tuning the gain and bias term of the LayerNorm module with only 0.03\% parameters. As we continue to research the unified framework of combining LN-tuning with earlier techniques, we discover that: (1) SOTA performance is achieved by the unified framework of combining prefix-tuning, which is one of the adapter-based techniques using MHA, and LN-tuning. (2) While unified frameworks that tune MHA and LayerNorm simultaneously can increase performance, those that simultaneously tune FFN and LayerNorm will have the opposite effect. Further, LN-tuning is better understood by an ablation investigation and a visualization experiment of the bias and gain terms.
		
	\end{abstract}
	\vspace{-0.2cm}
	
	\section{Introduction}
	\label{sect:intro}
	Natural language processing (NLP) is presently dominated by the transfer learning from Pre-trained Language Models (PLMs) paradigm~\cite{BERT, PLM-2}, which produces superior results in many tasks ~\cite{PLMs_Survey, ELMo, BERT}. The typical method used by PLMs to integrate the information they gained during the pre-training stage into downstream tasks is fine-tuning. A copy of the model needs to be retrained and saved for each downstream operation, which could be expensive given the enormous size of modern PLMs. To address the aforementioned issue, parameter-efficient tuning techniques have been proposed, which only modify a small subset of the pre-trained parameters and freeze the majority of them. To make measurable progress in this area, a lot of work has been done. ~\citeauthor{Encoder-Agnostic-Adaptation, Houlsby-Adapter, AdapterFusion, Adapter-Eff} propose several adapter techniques that insert trainable bottleneck layers, i.e., learnable down and up projections into the Feed-forward Network layer of each PLM block. Prefix-tuning~\cite{Prefix-Tuning}, P-tuning v2~\cite{deep-prompt-tuning}, and deep prompt tuning are used in MHA to optimize MLP networks and achieve continuous prefix prompt. More recently, research efforts have been made to create a unified framework that simultaneously tunes the representations of MHA and FFN, including those of the MAM adapter~\cite{MAM} and UniPELT~\cite{UniPELT}. Prefix-tuning is essentially a form of adapter-based method that is effective in MHA, as noted by MAM adapter. By integrating adapter-based approaches that operate on both MHA and FFN, they are able to attain SOTA performance. It is clear from this that earlier approaches in this area included tunable adapters to the MHA or/and FFN of Transformer blocks to provide parameter-efficient tuning. Nevertheless, the power of LayerNorm for parameter-efficient tuning is overlooked while being a crucial component of Transformer-based PLMs. Following the normalization of mean and variance, the gain and bias terms are applied for affine transformation on each input neuron in LayerNorm, acting as a fine-grained adaptive module on the data~\cite{layernorm_geo}. In earlier techniques, these modules are trained on a sizable general corpus and unsupervised tasks but left unchanged when adapted into a downstream dataset of a particular domain and supervised tasks, causing an unreasonable transformation since the gap from both the data and the task is quite large between pre-training and fine-tuning stage. In this research, we provide a straightforward but efficient technique called LN-tuning with the learnable gain and bias term of LayerNorm to close the aforementioned gaps. Following are some examples of our contribution: 

\begin{itemize}

\item We propose LN-tuning, which first explores the potential of LayerNorm for parameter-efficient tuning, achieving comparable performance to prior approaches with a very small number of parameters and a very high time efficiency. 

\item Prefix-tuning combined with LN-tuning leads to SOTA performance, outperforming MAM, the adapter-based unified framework that tunes MHA and FFN simultaneously. 

\item We empirically discover that while tuning both MHA and LayerNorm inside a single framework can enhance performance, doing the same with FFN and LayerNorm can actually worsen it. 

\item LN-tuning is better understood thanks to the ablation study of terms, layers, and modules, as well as the visualization experiment of the gain and bias term.
	
\end{itemize}
 
	\section{Preliminaries: Parameter-Efficient Tuning}
	\label{sect:preli}
	\subsection{Existing Approaches}

 The common parameter-efficient tuning techniques fall into three categories: those only tunes FFN of PLMs, those only tunes MHA of PLMs, those tunes MHA and FFN simultaneously in a unified framework. 
	
    Parallel Adapters insert a trainable bottleneck after FFN of each Transformer block, which is indeed a special FFN with down projection and up projection and open the magic box of parameter-efficient tunning for PLMs. Further researches~\cite{AdapterFusion, AdapterHub} studied variants of adapters in different positions. Among them, Scaled Sequential Adapter~\cite{MAM} is shown to have the best performance empirically. 

    Prefix-tuning~\cite{Prefix-Tuning} and P-tuning v2~\cite{P-Tuning-v2} are proposed to optimize additional past key values in the MHA of each Transformer block, which work as virtual tokens to supply task-specific prompts for PLMs. By learning the whole prefix net in training and only saving past key values obtained from this net for inference, prompt tuning goes a notable step further for parameter-efficient tuning by working in MHA.  

    The recent unified works, MAM adapter and UniPELT are proposed to combine prefix-tuning and adapter module, tuning the MHA and FFN of each Transformer block simultaneously. Those unified frameworks show empirical improvement compared with non-unified methods, with more tunable parameters and more infeasible tunable layers. 
    
    In addition, there are also several other parameter-efficient methods. BitFit only tunes bias vectors of Transformer parameters. Diff-pruning~\cite{Diff} learns a sparse parameter update vector.~\citeauthor{3V} use three learnable vectors to achieve transferability for Natural Language Understanding tasks, which we name 3V
    for brevity of reference.

    \subsection{Discusion} 
 By tuning MHA and/or FFN of PLMs, earlier techniques in this field were able to produce parameter-efficient results. However, as a key component of Transformer-based PLMs, the potential of LayerNorm for parameter-efficient tuning is ignored. Is it possible to apply a new tuning technique to PLMs' LayerNorm by tuning a few parameters? Only unfreezing its own parameters while keeping other PLM parameters frozen can function properly since each LayerNorm module has inherent scale and shift operations. This is what we refer to as LN-tuning.

	
	\section{Method}
	\subsection{LN-Tuning}
	\label{sect:method}
    Layer normalization (LayerNorm) is a technique to normalize the distributions
of intermediate layers. It enables smoother gradients, faster training, and better
generalization accuracy~\cite{AdaNorm}.
	As Eq.~\ref{eq:LN} shows, LayerNorm involves two stages: (1) normalize $\bm{x}$ by mean and variance (2) forward by the scale and shift operations consisting of the gain term $\bm{g}$  and bias term $\bm{b}$, respectively.
	
	Our proposed LN-Tuning keeps parameters in the gain term (for scale operation) and bias term (for shift operation) trainable, which are initialized from the pre-training stage, while fixing other parameters of PLMs. 
	
	\begin{equation}
		\label{eq:LN}
		\begin{aligned}
		\mathrm{LayerNorm}(\bm{x}) = \frac{\bm{g}}{\sigma} \odot (\bm{x} - \mu) + \bm{b} 
		\end{aligned}
	\end{equation}
	
	where 
       	\begin{equation} \nonumber
		\mu = \frac{1}{H}\sum_{i=1}^{H}{\bm{x}_i}~~~~~~
		\sigma = \sqrt{\frac{1}{H}\sum_{i=1}^{H}{(\bm{x}_i - \mu)^2}}
	\end{equation}

	\begin{figure}[t!]
	\centering
			\includegraphics[width=0.42\textwidth]{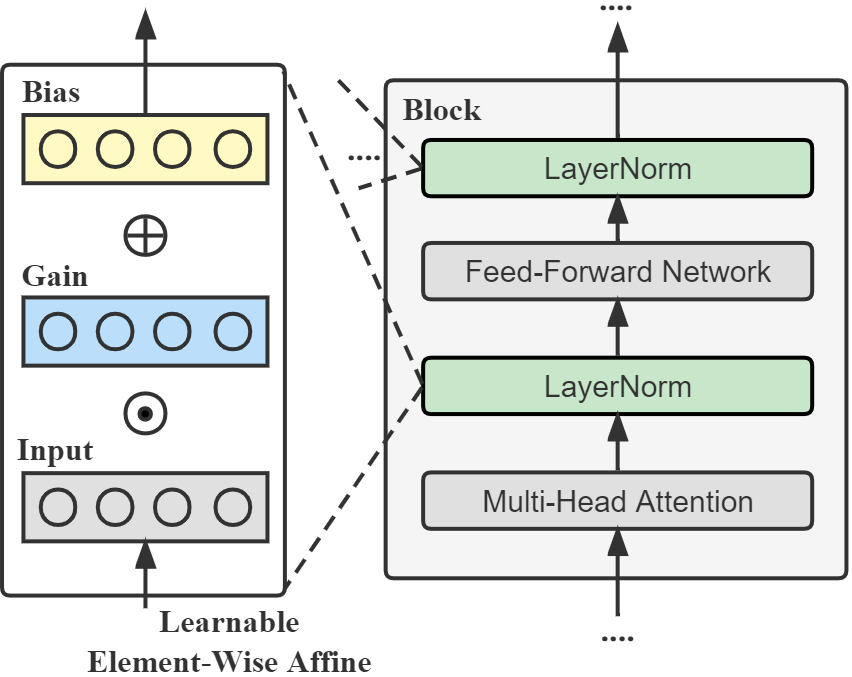} 	 
	\caption{Illustration of our proposed LN-tuning.}
	\label{fig:time}	
\end{figure}

	\subsection{Combine LN-Tuning and Previous Methods}
     LN-tuning can also be combined with other parameter-efficient to work as a unified framework. We explore five methods:
	
	(1)~Scaled Parallel Adapter with FFN + LN-tuning \\
	(2)~Sequential Adapter after FFN + LN-tuning \\
	(3)~Sequential Adapter after MHA + LN-tuning \\
	(4)~Prefix-tuning\footnote{As Sect.~\ref{sect:intro} illustrates, MAM Adapter~\cite{MAM} points out that prefix-tuning is equal to Scaled Parallel Adapter with MHA.} + LN-tuning \\
	(5)~BitFit + LN-tuning
	
	More details about four different adapter-based variants, i.e., the above methods (1) to (4), can be found in appendix~\ref{appendix:mha_ffn}.
	\label{combine_intro}

	\section{Experiments}
	\label{sect:exper}
	We validate the effectiveness of the proposed method on 11 benchmark datasets and seven types of downstream tasks, including both NLU and NLG ones, with the presence of six state-of-the-art baselines.   
	\subsection{General Setup}
	\label{ssect:setup}
	
	\textbf{Task Setup}.~To evaluate the proposed LN-tuning comprehensively, we conduct \emph{cross-task},
	\emph{cross-PLM-architecture}, and \emph{cross-PLM-scale} experiments. For cross-task validation, we conduct both NLU and NLG tasks. Specifically, for NLU tasks,
	we choose seven type datasets: \textbf{(1) Named-Entity Recognization (NER)}, including CoNLL2004~\cite{conll2004} and Twitter~\cite{twitter}; \textbf{(2) Natural Language Inference (NLI)}, including SNLI~\cite{SNLI} and CB~\cite{SuperGLUE};
	\textbf{(3) Paraphrase Identification (PI)}, including  SICK~\cite{SICK};
	\textbf{(4) Sentiment Analysis (SA)}, including SST-2~\cite{GLUE}; 
	\textbf{(5) Question Answering (QA)}, including CSQA~\cite{CSQA} and SocIQA~\cite{SocIQA}; \textbf{(6) Table-to-Text Generation}, including E2E~\cite{E2E} and DART~\cite{DART}; \textbf{(7) Dialogue Summarization}, including Samsum~\cite{Samsum} .
	
	The cross-PLM-architecture validation requires approaches
	to be verified on both encoder-only (BERT~\cite{BERT}) and decoder-only (GPT-2~\cite{GPT-2}) Transformer architecture. The cross-PLM-scale validation requires approaches to be verified on PLMs of different scales. Specifically, the same experiments for NLU are conducted on both $\text{BERT}_{\text{base}}$~\footnote{\href{https://huggingface.co/bert-base-uncased}{\texttt{https://huggingface.co/
	bert-base-uncased}}} and $\text{BERT}_{\text{large}}$~\footnote{\href{https://huggingface.co/bert-large-uncased}{\texttt{https://huggingface.co/
	bert-large-uncased}}},  
	while $\text{GPT-2}_{\text{medium}}$~\footnote{\href{https://huggingface.co/gpt2-medium}{\texttt{https://huggingface.co/gpt2-medium}}} is for NLG.
	
	\textbf{Baseline Methods}.~
    We compare our methods with six state-of-the-art tuning methods including full-tuning, scaled parallel adapter-tuning~\cite{AdapterFusion, MAM}, prefix-tuning~\cite{P-Tuning-v2}, MAM adapter~\cite{MAM},  BitFit~\cite{Bitfit} and 3V\footnote{Since it uses three trainable vectors to achieve parameter-efficient tuning and the orginal paper didn't name their method, we name it 3V in our paper for clarity and brevity. Also, we carefully write the code of this method ourselves according to the details described in original paper because they haven't release their code yet.}~\cite{3V}. For brevity,3 we agree to use adapter, prefix, MAM to represent scaled parallel adapter-tuning, prefix-tuning, and MAM Adapter respectively in all tables of this paper.

    We align the tunable amount of additional parameters of different methods to ensure a fair comparison, which is accomplished by setting hyperparameters. Specifically, for prefix-tuning, the hyperparameter to be adjusted is its prefix length $l$, where
    we set $l=16$ for $\text{BERT}_{\text{base}}$, $l=24$ for $\text{BERT}_{\text{large}}$, and $l=16$ for $\text{GPT-2}_{\text{medium}}$. For adapter, we adjust the intermediate dimension $d_b$, where
    we set $d_b=16$ for $\text{BERT}_{\text{base}}$, $d_b=24$ for $\text{BERT}_{\text{large}}$, $d_b=16$ for $\text{GPT-2}_{\text{medium}}$.  For MAM adapter, we adjust the both, keeping $d_b=l=8$ for $\text{BERT}_{\text{base}}$, 
    $d_b=16, l=8$ for $\text{BERT}_{\text{large}}$,  and 
    $d_b=8, l=8$ for $\text{GPT-2}_{\text{medium}}$.

    \textbf{Implementation Details}.~We conduct experiments on two NVIDIA GeForce RTX 3090 GPUs. The results are evaluated by different measures as suggested by different tasks. To reduce the interference of randomness, we repeat the experiments for three times and the average scores (for NLU) or the rank (for NLG) is returned as results. According to the recorded experience~\cite{Houlsby-Adapter, AdapterHub, Prefix-Tuning, MAM}, the common hyper-parameters are adjusted according to the statistical characteristics of datasets. 
	
	For NLU tasks, we set the training epoch 30, with an early stopping strategy of 10 non-decrease validation loss. The batch size setting can be found in Table~\ref{table:bs_nlu} of Appendix \ref{appendix:bs}. For LN-tuning, we adjust the learning rate from the priority order in \{1e-2, 1e-3, 2e-4\}~\footnote{We empirically find that LN-tuning needs larger learning rate than other approaches in some datasets.}. We adjust the learning rate from the priority order in \{1e-3, 2e-4\} for other methods.
	
	For NLG tasks, we set the training epoch 20. The batch size setting can be found in Table~\ref{table:bs_nlg}  of Appendix \ref{appendix:bs}, and the learning rate is 2e-4 for all methods. The E2E dataset contains about 50K examples whose average output length is 22.9. We use the official evaluation script~\footnote{\href{https://github.com/tuetschek/e2e-metrics}{\texttt{https://github.com/tuetschek/
	e2e-metrics}}} to calculate BLEU~\cite{BLEU}, NIST~\cite{NIST}, METEOR~\cite{METEOR}, ROUGE-L~\cite{ROUGE}, and CIDEr~\cite{CIDEr}. The Samsum dataset contains about 15K examples, whose average output length is 23.7. We use the standard python package \texttt{rouge} to calculate ROUGE-1, ROUGE-2, ROUGE-L~\cite{ROUGE}. The DART dataset consists of 82K examples, whose average output length is 27.3. We use the official evaluation script~\footnote{\href{https://github.com/Yale-LILY/dart}{\texttt{https://github.com/Yale-LILY/dart}}} to calculate BLEU, METEOR, and TER~\cite{TER}. We use $\text{GPT-2}_{\text{medium}}$~\cite{GPT-2}
	as the experimental PLM, where the max generation length is set to [35, 35, 45] for [E2E, Samsum, DART], respectively.

\subsection{Main Results}
  
\begin{table*}[ht]
\centering
	\resizebox{0.92 \textwidth}{!}{
\begin{tabular}{@{}ccccccccccc@{}}
\toprule
Method  & \#Para.  & CN04          & Twiiter       & SICK          & SNLI          & SST-2         & CB            & CSQA          & SocIQA        & Avg.          \\ \midrule
\multicolumn{11}{c}{BERT-Large}                                                                                                                                    \\
FT      & 100\%    & \textbf{85.2} & 75.8          & 86.2          & \textbf{85.4} & 92.8          & \textbf{80.4} & \textbf{69.8} & 63.4          & \textbf{79.9} \\
Adapter & 0.33\%   & 82.8          & 76.3          & {\ul 86.4}    & 85            & {\ul 93.0}      & 74.1          & 62.6          & 65.3          & 78.2          \\
Prefix  & 0.33\%   & 81.4          & 76.2          & 86.3          & {\ul 85.3}    & \textbf{93.4} & 75.0            & {\ul 63.2}    & {\ul 65.4}    & 78.3          \\
MAM     & 0.66\%   & {\ul 83.0}      & \textbf{78.1} & \textbf{86.6} & 85.2          & 93.1          & {\ul 77.6}    & {\ul 63.2}    & \textbf{65.5} & {\ul 79.0}      \\
3V      & 0.0006\% & 68.1          & 73.6          & 81.3          & 82.8          & 89.1          & 70.2          & -             & -             & -             \\
BitFit  & 0.07\%   & 79.2          & 74.2          & 77.8          & 81.6          & 92.6          & 70.5          & 59.7          & 62.8          & 74.8          \\ \midrule
LN      & 0.03\%   & 78.9          & {\ul 76.9}    & 85.8          & 83.8          & 89.8          & 70.5          & 59.6          & 63.3          & 76.1          \\ \midrule
\multicolumn{11}{c}{BERT-Base}                                                                                                                                     \\
FT      & 100\%    & \textbf{87.2} & 75.3          & 84.5          & 84.2          & 90.9          & \textbf{82.7} & 50.2          & 55.0            & 76.3          \\
Adapter & 0.28\%   & 72.5          & 75.7          & 83.7          & {\ul 84.4}    & 91.5          & {\ul 73.8}    & \textbf{60.6} & {\ul 61.6}    & 75.5          \\
Prefix  & 0.28\%   & 77.9          & 75.9          & 84.2          & 84.0            & \textbf{91.9} & 76.8          & {\ul 60.4}    & {\ul 61.6}    & {\ul 76.6}    \\
MAM     & 0.56\%   & {\ul 80.3}    & {\ul 76.3}    & {\ul 84.8}    & \textbf{84.5} & {\ul 91.6}    & {\ul 73.8}    & {\ul 60.4}    & \textbf{61.8} & \textbf{76.7} \\
3V      & 0.0014\% & 67.2          & 70.7          & \textbf{85.0}   & 82.2          & 88.1          & 72.0            & -             & -             & -             \\
BitFit  & 0.08\%   & 80.9          & 71.5          & 74.4          & 79.9          & 89.9          & 68.5          & 55.3          & 57.6          & 72.2          \\ \midrule
LN      & 0.04\%   & 79.1          & \textbf{76.7} & 74.0            & 82.4          & 91.4          & {\ul 73.8}    & 58.5          & 58.8          & 74.3          \\ \bottomrule
\end{tabular}}

	\caption{Results with $\text{BERT}_{\text{large}}$ and $\text{BERT}_{\text{base}}$. We report the average score with the standard deviation as the subscript. The \textbf{best} and \underline{2nd best} methods on each dataset are in bold and underlined, respectively.*3V can not be applied into these two QA tasks and thus is omitted to calculate the average values and rank metric. }
	\label{nlu_main}

\end{table*}
  
\begin{table*}[]
\resizebox{1 \textwidth}{!}{%
\begin{tabular}{@{}cc|ccccc|ccc|cccccc|c@{}}
\toprule
\multirow{2}{*}{Method} & \multirow{2}{*}{\#Para.} & \multicolumn{5}{c|}{E2E}                                                         & \multicolumn{3}{c|}{Samsum}                      & \multicolumn{6}{c|}{WebNLG}                                                                    & \multirow{2}{*}{Rank} \\
                        &                          & BLEU           & NIST          & MET            & R-L            & CIDEr         & R-1            & R-2            & R-L            & BLEU           & MET           & TER↓          & Mover         & BERT          & BLEURT        &                       \\ \midrule
FT                      & 100\%                    & {\ul 65.07}    & \textbf{8.61} & 43.42          & 67.90          & {\ul 2.38}    & \textbf{44.70} & \textbf{20.37} & \textbf{41.57} & \textbf{39.43} & \textbf{0.34} & {\ul 0.55}    & \textbf{0.65} & \textbf{0.93} & \textbf{0.39} & \textbf{2.68}         \\
Adapter                 & 0.13\%                   & 64.93          & 8.46          & \textbf{44.21} & {\ul 68.63}    & \textbf{2.39} & 43.23          & 18.67          & 40.17          & 38.40          & {\ul 0.33}    & 0.56          & 0.64          & \textbf{0.93} & {\ul 0.38}    & 4.12                  \\
Prefix                  & 0.13\%                   & \textbf{65.27} & {\ul 8.55}    & {\ul 43.70}    & 68.27          & 2.37          & {\ul 43.70}    & {\ul 19.97}    & {\ul 40.83}    & {\ul 38.87}    & {\ul 0.33}    & \textbf{0.54} & \textbf{0.65} & \textbf{0.93} & {\ul 0.38}    & {\ul 3.30}             \\
MAM                     & 0.26\%                   & 64.80           & 8.46          & 43.90          & \textbf{68.67} & 2.36          & 43.50          & 19.40          & 40.33          & {\ul 38.87}    & {\ul 0.33}    & {\ul 0.55}    & \textbf{0.65} & \textbf{0.93} & {\ul 0.38}    & 3.63                  \\
BitFit                  & 0.09\%                   & 64.27          & 8.54          & 41.80          & 67.63          & 2.17          & 39.27          & 15.23          & 36.17          & 35.33          & 0.30           & 0.61          & 0.62          & 0.92          & 0.32          & 6.56                  \\ \midrule
LN                      & 0.03\%                   & 64.07          & 8.34          & 43.63          & 67.97          & 2.35          & 42.77          & 18.80          & 39.53          & 38.47          & {\ul 0.33}    & {\ul 0.55}    & 0.64          & \textbf{0.93} & 0.36          & 4.25                  \\ \bottomrule
\end{tabular}}

\vspace{11pt}

	\caption{Results with $\text{GPT-2}_{\text{medium}}$. We report the average score with the standard deviation as the subscript. The \textbf{best} and \underline{2nd best} methods on each dataset are in bold and underlined respectively. 3V can not be applied into NLG tasks and thus is omitted as a baseline here. }
	\label{nlg_main}
\end{table*}
 
In Table~\ref{nlu_main}, we present the comparison results for the NLU tasks on $\text{BERT}_\text{large}$ and $\text{BERT}_\text{base}$. It is clear from this that full-tuning and MAM adapter may typically achieve superior performance. Better performance is expected because more recently introduced parameters and multiple PLM modules need to be tuned. Compared to other earlier approaches, 3V and BitFit performs the poorest with less parameters. Under the tunable parameter alignment setting, the performance of prefix tuning and adapter tuning is comparable to one another. 

The performance of LN-tuning is then examined. Comparing approaches whereas the ratio of the tunable parameters is more significant than 0.3\%, LN-tuning is inferior to them by tuning only 0.03\%–0.04\% of parameters. By using almost half the tunable parameters of BitFit, LN-tuning performs much better than BitFit. LN-tuning outperforms 3V in terms of performance and is also applicable to a wider variety of NLP tasks than 3V, including QA tasks for NLU and NLG tasks.

With a few limited differences, the methods' overall performance in NLG tasks is similar to that in NLU tasks. First, prefix-tuning outperforms MAM adapter. Second, our LN-tuning exhibits a performance closer to that of adapter-based approaches, such as adapter and MAM adapter, compared to the NLU task.

	\subsection{Efficiency Analysis}
	
		\begin{figure*}[ht]
	\centering
	\subfigure[BERT-base Training]{
		\label{fig:base_train}
		\begin{minipage}[t]{0.31\linewidth}	
			\centering
			\includegraphics[width=1\textwidth]{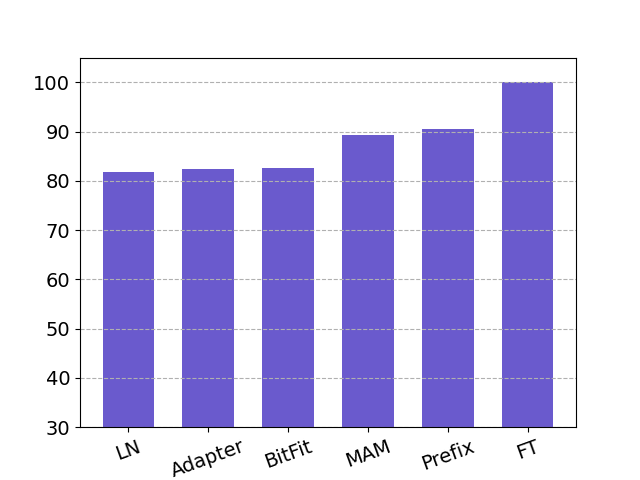} 	 
		\end{minipage}
	} 
	\subfigure[BERT-large Training]{
		\label{fig:large_train}
		\begin{minipage}[t]{0.31\linewidth}	
			\centering
			\includegraphics[width=1\textwidth]{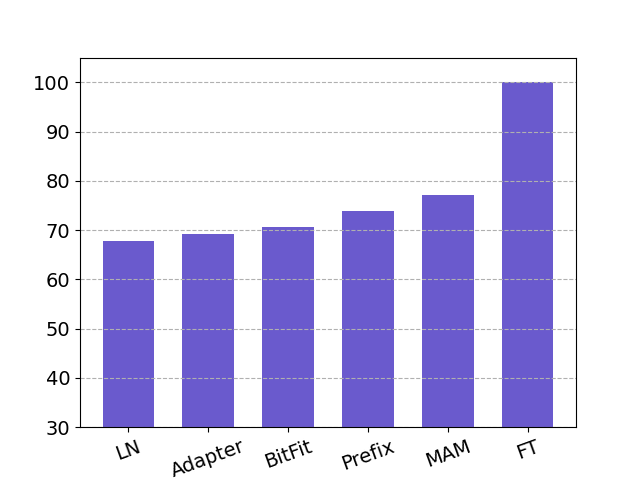}
		\end{minipage}
	} 
	\subfigure[GPT-2 Medium Training]{
		\label{fig:gpt_train}
		\begin{minipage}[t]{0.31\linewidth}	
			\centering
			\includegraphics[width=1\textwidth]{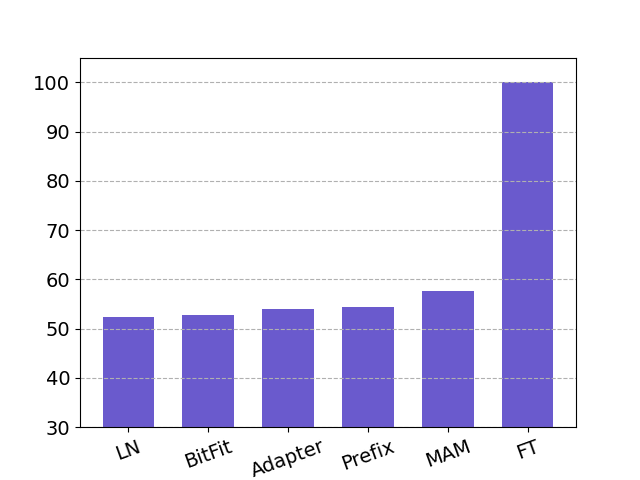}
		\end{minipage}
	} 
	\subfigure[BERT-base Inference]{
		\label{fig:gpt_infer}
		\begin{minipage}[t]{0.31\linewidth}	
			\centering
			\includegraphics[width=1\textwidth]{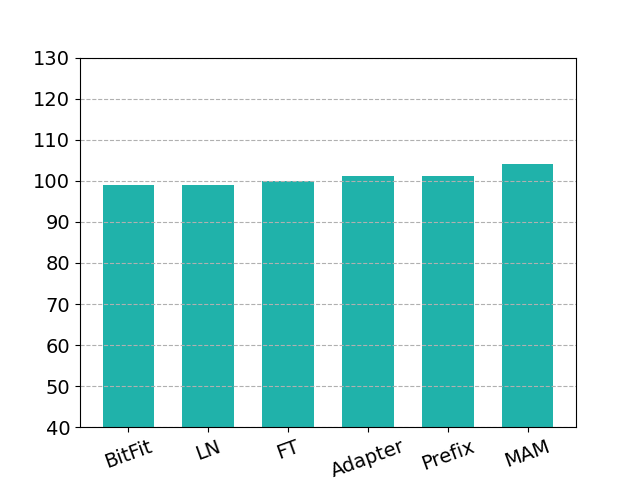}
		\end{minipage}
	} 
	\subfigure[BERT-large Inference]{
		\label{fig:large_infer}
		\begin{minipage}[t]{0.31\linewidth}	
			\centering
			\includegraphics[width=1\textwidth]{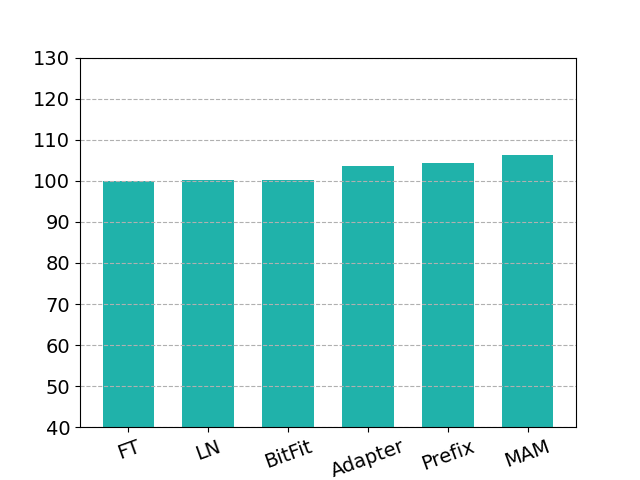}
		\end{minipage}
	} 
	\subfigure[GPT-2 Medium Inference]{
		\label{fig:gpt_infer}
		\begin{minipage}[t]{0.31\linewidth}	
			\centering
            \includegraphics[width=1\textwidth]{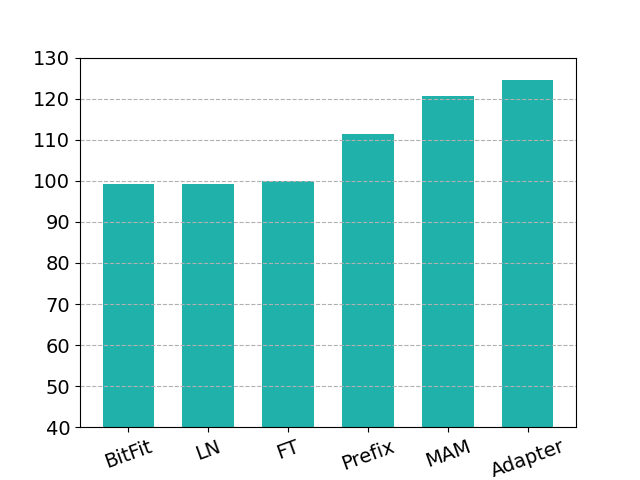}
		\end{minipage}
	}
	 
	\caption{Time Efficiency Comparison.}
	\label{fig:time}	
\end{figure*} 

 \vspace{11pt}
 
	\textbf{Setup}. In order to compare the training and inference time efficiency between our method and earlier ones, we generate statistics from running logs. Then, in comparison to Full-Tuning (FT), we report the relative training and inference times. This includes the average time costs for three NLG datasets for GPT-2 and eight NLU datasets for BERT. The time cost of FT is normalized to 100. 

\textbf{Result}.~As shown in Fig.~\ref{fig:time}, our proposed LN-tuning takes the least time in all PLM architectures for training process. LN-tuning, along with BitFit and FT, costs the similar least time for inference process as expected. The above results on both training and inference show the significant superiority of our method in time efficiency comparing previous adapter-based methods. 

In Fig.\ref{fig:base_train}, all parameter-efficient methods require training times that are less than 90\% of those of FT in $\text{BERT}_{\text{base}}$ and less than 80\% of those of FT in $\text{BERT}_{\text{large}}$, demonstrating that parameter-efficient methods can train PLMs of greater scales more quickly. From Fig.~\ref{fig:base_train}, Fig.~\ref{fig:large_train} and Fig.~\ref{fig:gpt_train}, we can observe that parameter-efficient methods show higer time efficiency in training in NLG tasks than in NLU tasks comparing with FT. However, whether in training or inference, MAM adapter typically has the lowest time efficiency, demonstrating that the unified methods of both tuning MHA and FFN require a significant investment in computational resources despite being able to produce better performance. Further, adapter-tuning shows higher time efficiency than prefix-tuning in training and inference, except for the NLG inference process.

\begin{table*}[!ht]
\centering
\resizebox{0.88 \textwidth}{!}{%
\begin{tabular}{@{}l|llllllll|l@{}}
\toprule
Method                       & CN4  & Twitter & SICK & SNLI & SST2 & CB   & CSQA & SociQA & Avg.          \\ \midrule
Scaled Parallel Adapter(FFN) & 82.8 & 76.3    & 86.4 & 85.0 & 93.0 & 74.1 & 62.6 & 65.3   & \textbf{78.2} \\
+LN                          & 82.3 & 77.6    & 86.0 & 84.8 & 93.5 & 72.3 & 61.3 & 65.0   & 77.8          \\
\midrule
Sequential Adapter(FFN)      & 76.0 & 76.8    & 85.1 & 82.8 & 93.1 & 71.4 & 62.2 & 62.8   & \textbf{76.2} \\
+LN                          & 74.7 & 76.8    & 84.1 & 82.6 & 92.9 & 72.6 & 62.2 & 62.4   & 76.0          \\ \midrule
Sequential Adapter(MHA)      & 76.1 & 71.9    & 80.0 & 81.7 & 88.8 & 71.4 & 60.8 & 61.0   & 74.0          \\
+LN                          & 77.8 & 74.8    & 81.4 & 82.9 & 92.6 & 74.1 & 62.0 & 62.8   & \textbf{76.0} \\ \midrule
Prefix                       & 81.4 & 76.2    & 86.3 & 85.3 & 93.4 & 75.0 & 63.2 & 65.4   & 78.3          \\
+LN                          & 84.2 & 77.2    & 86.6 & 85.4 & 93.8 & 81.2 & 64.0 & 65.5   & \textbf{79.8} \\
\midrule
MAM +LN                          & 81.4 & 77.4    & 86.0 & 85.0 & 93.1 & 80.4 & 63.2 & 65.4   & \textbf{79.0} \\ \midrule
BitFit                       & 79.2 & 74.2    & 77.8 & 81.6 & 92.6 & 70.5 & 59.7 & 62.8   & 74.8          \\
+LN                          & 79.1 & 74.1    & 85.4 & 83.7 & 92.9 & 71.4 & 62.7 & 63.6   & \textbf{76.8} \\ \bottomrule
\end{tabular}}
	\caption{Results of comparison between five previous parameter-efficient methods and the unified framework of combining them with LN-tuning respectively. It is conducted on $\text{BERT}_{\text{large}}$.}
	\label{unified}
\end{table*}

\begin{table*}[!ht]
\centering
\resizebox{0.89 \textwidth}{!}{%
\begin{tabular}{@{}ccccccccccc@{}}
\toprule
Ablation Type           & Method           & CN04          & Twitter       & SICK          & SNLI          & SST2          & CB            & CSQA          & SociQA        & Avg           \\ \midrule
\multicolumn{11}{c}{BERT-Large}                                                                                                                                                            \\
-                       & \textit{Full*}   & \textit{80.2} & \textit{77.2} & \textit{84.9} & \textit{84.0} & \textit{91.9} & \textit{74.1} & \textit{60.5} & \textit{63.2} & \textit{77.0} \\
\multirow{2}{*}{Term}   & Only Gain        & 69.5          & 69.5          & 76.3          & 80.9          & 91.6          & 71.4          & 53.3          & 57.9          & 71.3          \\
                        & Only Bias        & 79.8          & 72.6          & 77.0          & 81.2          & 91.8          & 73.2          & 55.8          & 60.9          & 74.0          \\
\multirow{2}{*}{Module} & Only FFN         & 77.3          & 76.5          & 82.2          & 81.9          & 92.6          & 72.8          & 55.6          & 61.0          & \textbf{75.0} \\
                        & Only MHA         & 75.8          & 77.4          & 82.0          & 81.6          & 92.2          & 72.3          & 56.2          & 58.8          & {\ul 74.6}    \\
\multirow{2}{*}{Layer}  & Only Layer 1-12   & 73.2          & 75.1          & 82.4          & 78.4          & 91.8          & 73.1          & 51.7          & 56.4          & 72.8          \\
                        & Only Layer 13-24  & 73.8          & 75.7          & 82.4          & 78.6          & 93.2          & 72.9          & 53.8          & 56.6          & 73.4          \\ \midrule
\multicolumn{11}{c}{BERT-Base}                                                                                                                                                             \\
-                       & \textit{Full*}   & \textit{79.8} & \textit{76.4} & \textit{81.0} & \textit{83.3} & \textit{91.4} & \textit{70.2} & \textit{57.9} & \textit{59.1} & \textit{74.9} \\
\multirow{2}{*}{Term}   & Only Gain        & 72.9          & 68.8          & 67.5          & 76.7          & 87.7          & 73.2          & 50.0            & 52.9          & 68.7          \\
                        & Only Bias        & 76.5          & 67.8          & 77.5          & 76.3          & 89.7          & 71.4          & 51.1          & 53.4          & 70.5          \\
\multirow{2}{*}{Module} & Only FFN         & 79.1          & 76.6          & 81.5          & 77.0          & 91.6          & 76.2          & 53.3          & 53.8          & \textbf{73.6} \\
                        & Only MHA         & 78.4          & 76.5          & 81.8          & 77.2          & 91.2          & 75.0          & 52.6          & 54.0          & {\ul 73.3}    \\
\multirow{2}{*}{Layer}  & Only Layer 1-6  & 78.2          & 76.0          & 67.9          & 74.1          & 90.7          & 74.4          & 50.8          & 50.6          & 70.3          \\
                        & Only Layer 7-12 & 71.3          & 74.9          & 68.2          & 73.9          & 90.8          & 73.8          & 50.3          & 50.3          & 69.2          \\ \bottomrule
\end{tabular}}

	\caption{Results of ablation study about terms, layers and modules with $\text{BERT}_{\text{large}}$ and $\text{BERT}_{\text{base}}$}. *We use 
 italic font to show results of the full LN-tuning, which is as a standard for comparison.
	\label{ablation}

\end{table*}

	\subsection{Unified Framework}
	\textbf{Setup}. Since LN-tuning operates in LayerNorm, the question of whether it may be combined with earlier parameter-efficient methods working in MHA or/and FFN to operate as a unified framework and achieve further performance improvement arises naturally. In this section, we explore the unified framework combing LN-Tuning with five approaches, as illustrated in Sec.~\ref{combine_intro}, which can be divided into four types according to tunable modules: (1) MHA only (2) FFN only (3) MHA and FFN simultaneously (4) BitFit. For the experiment, we employ the variable $\text{BERT}_\text{large}$. 
	 \label{sect:unified}

\textbf{Result}. As shown in Table~\ref{unified}, among Prefix+LN and Sequential Adapter (MHA) + LN, which are unified methods of combining methods working in MHA with LN-tuning, perform significantly better than the original single ones. Notably, the Prefix+LN even outperformed MAM Adapter, which is naturally Prefix + Scaled Parallel Adapter (FFN), in terms of SOTA performance. However, the unified methods which combine adapter-based methods working in FFN with LN-tuning get a performance decrease. Empirically, this can be concluded that LN-tuning can combine with adapter-based methods working in MHA to improve further but will decrease performance if combined with those working in FFN. This explains why the performance of the LN + MAM Adapter is so comparable to that of the MAM Adapter, which is the result of performance enhancement brought on by prefix + LN and performance degradation brought on by scaled parallel adapter (FFN) + LN. This is so because the Prefix + Scaled Parallel  Adapter (FFN) is essentially what the MAM Adapter is. Finally, the unified framework of BitFit + LN displays a performance boost, which highlights the advantages of our LN-tuning in high combination compatibility with earlier approaches to get better performance.

\subsection{Ablation Study}
	\textbf{Setup}.~To explore whether LN-tuning may be enhanced to be more parameter-efficient, we undertake an ablation study from three aspects: terms, modules, and layers. Specifically, for terms, we only keep one option of the gain or the bias term trainable. For layers, we keep vectors of LayerNorm of only the half layers close to input or output trainable, {i.e. from layer 1 to 12 or from 13 to 24}, if using $\text{BERT}_\text{large}$. The same way is for using $\text{BERT}_\text{base}$. For modules, since there are two LayerNorm modules in each block of Transformer, where one is after MHA and the other is after FFN, we keep vectors trainable of only one module in each Transformer block. We use both $\text{BERT}_\text{large}$ and $\text{BERT}_\text{base}$ for the experiment in this section. 

\textbf{Result}.~As shown in Table~\ref{ablation}, comparing to full LN-tuning method, all ablated techniques obtain a performance drop, which validates no extraneous components for LN-tuning. Further, the influence of layers seems more critical than that of modules due to a larger performance decrease comparing ablated layer methods and ablated module methods. For term ablation type, the method with only the bias term performs better than that with only the gain term, whether in $\text{BERT}_\text{large}$ or $\text{BERT}_\text{base}$, which indicates that the bias term plays a more critical role than the gain term in LN-tuning. The added MHA learnable module looks more relevant for module ablation type than the added FFN learnable module. For layer ablation type, the layers adjacent to input seems to be more importan than that close to output in $\text{BERT}_\text{base}$, however, the outcome is the opposite for $\text{BERT}_\text{large}$. This shows that the importance of layers is quite different in different size of PLMs in LN-tuning and those layers close to output can play a more significant role in larger size of PLMs.
			
	\begin{figure}[t!]
	\centering
			\includegraphics[width=0.47\textwidth]{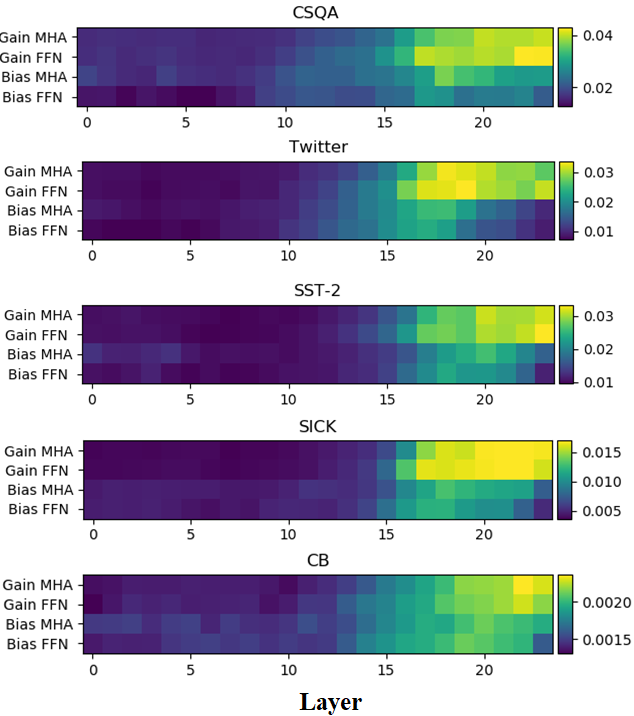} 	 
	\caption{Change in gain and bias term on five type of NLU tasks. `Gain MHA' means the gain term of LayerNorm module after MHA in each layer of PLMs, and so forth for other labels of Y-axis.}
	\label{fig:vis}	
\end{figure}

	\subsection{Visualization of Gain and Bias Term.}
	\textbf{Setup}.~We visualize the change of the gain and bias term on each layer of PLMs to give a further understanding about LN-tuning. Specifically, following BitFit, we use $\frac{1}{\text{dim}(\bm{t})}
	\Vert \bm{t}_o - \bm{t}_f \Vert_1$ to measure the amount of change for terms, where $\bm{t}$ represents the gain term $\bm{g}$ or the bias term $\bm{b}$ of LayerNorm, which means the average
    absolute change, across its dimensions, between the initial LM values $\bm{t}_o$ and its fine-tuned values $\bm{t}_f$. We choose five datasets which covers all type of NLU tasks in Sec.~\ref{ssect:setup} and use $\text{BERT}_\text{large}$ for the experiment.
	 
	\textbf{Result}.~As shown in Fig.~\ref{fig:vis}, there can be oberseved that the terms of layers close to the output, i.e. layer 15 to 24, changes more than those close to input, whether the gain or bias. Meanwhile, in those layers close to output, the gain term change more than bias term (This doesn't mean that the gain term is more important
 than the bias term in LN-tuning).
	Comparing results between tasks, the task complexity and the dataset scale may affect the extent of terms' change. Firstly, comparing SST-2 and the other two datasets of binary classification tasks, there is a greater change in terms of LN-tuning. This may be because that there are larger solution spaces (greater task complexity) for the QA (CSQA) and NER (Twitter) task than binary classification tasks such as sentiment analysis (SST-2), Paraphrase Identification (SICK) or Natural Language Inference (CB) task. There needs greater variation in the terms of LN-tuning in CSQA and Twitter dataset. Secondly, the order of term variation in binary classification tasks is SST-2 > SICK > CB, which is the same as the order of their data scale: SST-2 (67,349 items) > SICK (4,439 items) > CB (250 items). A reasonable explanation for this different degree of variation is that larger data sizes require a more significant term variation to accommodate a variety of data samples from a wider range of domains.

	\section{Related Work}
    \textbf{Parameter-Efficient Tuning for PLMs.} Given the large-scale size of current PLMs, it is infeasible to train and store complete copies of large PLMs for each downstream task. Parameter-efficient tuning methods are proposed to deal with it by efficiently tuning PLMs with few trainable parameters. Existing approaches include Adapter~\citep{Houlsby-Adapter, Adapter-Eff, AdapterFusion, AdapterHub} which tunes FFN only, prompt tuning~\citep{Prefix-Tuning, P-Tuning-v2} which tunes MHA only, Unified Frameworks\citep{MAM, UniPELT}, which tunes both simultaneously, and other solutions like BitFit~\citep{Bitfit} and Diff-pruning~\citep{Diff}. Our LN-Tuning, which can achieve comparable performance with fewer parameters and high time efficiency, first establishes the significance of LayerNorm for parameter-efficient tuning of PLMs in comparison to these parameter-efficient studies. Additionally, SOTA performance is attained by the combined framework of prefix-tuning and LN-tuning, which takes lightweight tuning of PLMs to a higher level of parameter efficiency.

	\textbf{Feed-forward Network.}
Our work also has a connection to the Feed-forward Network (FFN). The scale and shift operation in LN-tuning is a unique, sped-up FFN that only conducts projection on a single neuron, as opposed to linear aggregation between input layer neurons. By training a very small number of sets of parameters, our approach can efficiently transfer PLMs to downstream tasks. Previous research overlooked this streamlined FFN. By putting it to use in parameter-efficient tuning for PLMs, we are the first to discover its potent transferability.

	\section{Conclusions}
	\label{sect:concl}
 
In this paper, we first propose \emph{LN-Tuning}, which only tunes the bias and gain term of LayerNorm to enable parameter-efficient transferring for PLMs. Later, we investigate a unified framework for merging LN-tuning with earlier parameter-efficient techniques and discover that SOTA performance can be achieved by combining prefix-tuning with LN-tuning. We also draw the empirical conclusion that while tuning MHA and LayerNorm in a single framework will increase performance, doing the same for FFN and LayerNorm would result in worse performance. Finally, the ablation study of terms, layers, and modules, as well as the visualization experiment of the gain and bias term further understand LN-tuning.

	\bibliographystyle{acl_natbib}
	\bibliography{anthology}

 \clearpage
 
	\appendix
		\section{Adapter-based Variants.}
\begin{figure*}[t!]
	\centering
			\includegraphics[width=0.98\textwidth]{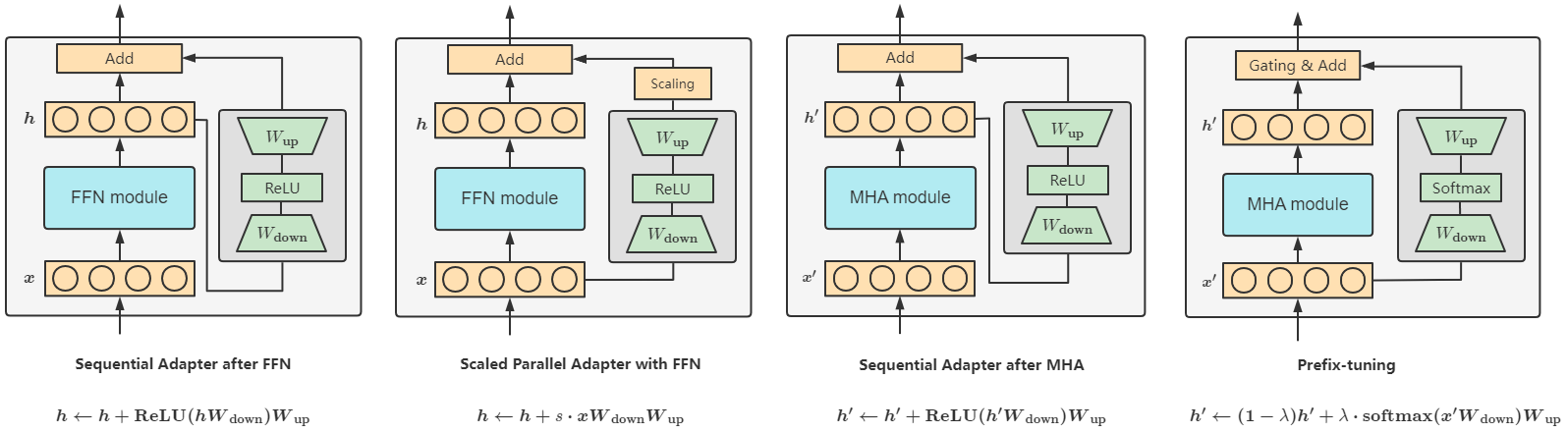} 	 
	\caption{Illustration of four types of adapter variants.}
	\label{fig:mha_ffn}	
\end{figure*}

Fig.~\ref{fig:mha_ffn} displays the four different adapter-based variants shown in Sect.~\ref{combine_intro} and used in Sect.~\ref{sect:unified}. The $\bm{W}_{\text{down}}$ and $\bm{W}_{\text{up}}$ matrices are tunable modules in each adapter.

		\label{appendix:mha_ffn}
		
	\section{Batch Size Settings}
 
The detailed batch size settings for NLU and NLG tasks are displayed in the Table~\ref{table:bs_nlu} and able~\ref{table:bs_nlg} respectively. In order to conduct a fair comparison, we make full use of the GPUs' VRAM capacity and work to make sure the batch size parameters for each approach are identical. We decrease the value of batch size to prevent a "CUDA Out of Memory" problem because full-tuning and MAM Adapter have more tunable parameters.
 
	\label{appendix:bs}
	 
\begin{table}[!ht]
\centering
\resizebox{0.5 \textwidth}{!}{
\begin{tabular}{@{}c|cccccccc@{}}
\toprule
Methods & CN04 & Twitter & SICK & SNLI & SST2 & CB & CSQA & SociQA \\ \midrule
        & \multicolumn{8}{c}{BERT-Base}                            \\
FT      & 128  & 128     & 512  & 512  & 256  & 48 & 48   & 48     \\
MAM     & 128  & 128     & 512  & 512  & 392  & 64 & 64   & 48     \\
Others  & 128  & 128     & 512  & 512  & 392  & 64 & 64   & 64     \\ \midrule
        & \multicolumn{8}{c}{BERT-Large}                           \\
FT      & 32   & 32      & 256  & 256  & 128  & 24 & 16   & 12     \\
MAM     & 48   & 48      & 256  & 256  & 256  & 32 & 24   & 24     \\
Others  & 48   & 48      & 256  & 256  & 256  & 32 & 32   & 24     \\ \bottomrule
\end{tabular}}
\caption{Batch size setting for NLU tasks.}
\label{table:bs_nlu}
\end{table}

\begin{table}[!ht]
\centering
 	\resizebox{0.25 \textwidth}{!}{
\begin{tabular}{@{}l|lll@{}}
\toprule
Method & Samsum & E2E & WebNLG \\ \midrule
FT     & 32     & 48  & 40     \\
Others & 36     & 96  & 84     \\ \bottomrule
\end{tabular}}
\caption{Batch size setting for NLG tasks.}
\label{table:bs_nlg}
 
\end{table}

\section{Supplementary Experiment of Prefix+LN}
\label{appendix:supply}
On both NLU and NLG tasks, we contrast the prior method for achieving SOTA performance with Prefix+LN. On NLG tasks, the unified prefix+LN technique outperforms prefix-tuning on most measures with an increase of just 0.03\% tunable parameters, as shown in Table~\ref{s1}. Prefix+LN outperforms MAM Adapter on NLU tasks in Table~\ref{s2}. This is true for both scales of BERT models with about just half tunable parameters, especially on $\text{BERT}_\text{large}$, which demonstrates the promise of prefix+LN for parameter-efficient tuning of large scale PLMs.

\begin{table*}[!ht]
\centering
\resizebox{1 \textwidth}{!}{ 
\begin{tabular}{@{}cc|ccccc|ccc|cccccc@{}}
\toprule
\multirow{2}{*}{Method} & \multirow{2}{*}{\#Para.} & \multicolumn{5}{c|}{E2E}                                                         & \multicolumn{3}{c|}{Samsum}                      & \multicolumn{6}{c}{WebNLG}                                                                     \\
                        &                          & BLEU           & NIST          & MET            & R-L            & CIDEr         & R-1            & R-2            & R-L            & BLEU           & MET           & TER↓          & Mover         & BERT          & BLEURT        \\ \midrule
Prefix                  & 0.13\%                   & \textbf{65.27} & 8.55          & 43.70          & 68.27          & 2.37          & 43.70          & 19.97          & 40.83          & 38.87          & 0.33          & \textbf{0.54} & \textbf{0.65} & \textbf{0.93} & \textbf{0.38} \\
Prefix+LN               & 0.16\%                   & 65.24          & \textbf{8.57} & \textbf{43.75} & \textbf{68.43} & \textbf{2.39} & \textbf{43.88} & \textbf{20.03} & \textbf{41.07} & \textbf{39.16} & \textbf{0.34} & \textbf{0.54} & \textbf{0.65} & \textbf{0.93} & \textbf{0.38} \\ \bottomrule
\end{tabular}}
\caption{The performance comparison between Prefix-tuning and Prefix-tuning+LN-tuning on NLG tasks. Even thought prefix-tuning achieve the best performance on NLU tasks over methods shown in Table~\ref{nlu_main}, our prefix-tuning+LN-tuning can obtain further performance improvement than prefix-tuning only.}
\label{s1}
\end{table*}

\begin{table*}[!ht]
\centering
\resizebox{0.85 \textwidth}{!}{ 
\begin{tabular}{@{}ccccccccccc@{}}
\toprule
Method    & \#Para. & CN04          & Twiiter       & SICK          & SNLI          & SST-2         & CB            & CSQA          & SocIQA        & Avg.          \\ \midrule
\multicolumn{11}{c}{BERT-Large}                                                                                                                                     \\
MAM       & 0.66\%  & 83.0            & \textbf{78.1} & \textbf{86.6} & 85.2          & 93.1          & 77.6          & 63.2          & \textbf{65.5} & 79            \\
Prefix+LN & 0.36\%  & \textbf{84.2} & 77.2          & \textbf{86.6} & \textbf{85.4} & \textbf{93.8} & \textbf{81.2} & \textbf{64.0} & \textbf{65.5} & \textbf{79.8} \\ \midrule
\multicolumn{11}{c}{BERT-Base}                                                                                                                                      \\
MAM       & 0.56\%  & 80.3          & \textbf{76.3} & \textbf{84.8} & 84.5          & 91.6          & 73.8          & 60.4          & \textbf{61.8} & 76.7          \\
Prefix+LN & 0.32\%  & \textbf{80.7} & 76.1          & 84.5          & \textbf{84.6} & \textbf{91.9} & \textbf{74.1} & \textbf{60.6} & 61.7          & \textbf{76.8} \\ \bottomrule
\end{tabular}}
\caption{The performance comparison between MAM Adapter and Prefix-tuning+LN-tuning on NLU tasks. Even thought MAM achieves the best performance on NLU tasks over methods shown in Table~\ref{nlg_main}, our prefix-tuning+LN-tuning can still outperform it.}
\label{s2}
\end{table*}

	\section{Future Work} 
While prefix-tuning and LN-tuning operate together to attain SOTA performance and LN-tuning has a high time efficiency with very few tunable parameters, there are still worthwhile areas for additional research. First, take note that the LN-tuning approach for tuning gain and bias term is a novel tuning technique that can be used after any PLM output vector. Exist any undiscovered techniques to perform SOTA by only learnable modules in LN-tuning? Further investigation can be done in future work to determine why the unified framework of integrating LN-tuning and Prefix-tuning (MHA+LN) can perform better than earlier adapter-based techniques (MHA+FFN).
     

\end{document}